\pdfoutput=1

\documentclass[11pt]{article}

\usepackage[]{EMNLP2022}

\usepackage{times}
\usepackage{latexsym}

\usepackage[T1]{fontenc}

\usepackage[utf8]{inputenc}

\usepackage{microtype}

\usepackage{inconsolata}

\usepackage{arydshln}
\usepackage{enumitem}
\usepackage{xcolor, soul}
\usepackage{graphicx}
\usepackage{multirow}
\usepackage{svg}
\usepackage{caption}
\usepackage{subcaption}
\usepackage{verbatim}
\usepackage{amsmath}
\usepackage{amssymb}
\usepackage{inconsolata}

\usepackage{balance}

\definecolor{OliveGreen}{rgb}{0,0.6,0}

\graphicspath{{img/}}

\newcommand{\shortarrow}[1][4pt]{\mathrel{%
   \hbox{\rule[\dimexpr\fontdimen22\textfont2-.2pt\relax]{#1}{1pt}}%
   \mkern-4mu\hbox{\usefont{U}{lasy}{m}{n}\symbol{41}}}}

\makeatletter
\def\@eqnnum{{\normalfont \color{red} (\theequation)}}
\makeatother

\title{Checks and Strategies for Enabling Code-Switched Machine Translation}

\author{Thamme Gowda \and Mozhdeh Gheini \and Jonathan May\\
  Information Sciences Institute \and Computer Science Department \\
  University of Southern California \\
  \texttt{\{tg,gheini,jonmay\}@isi.edu}
  }

\begin{document}
\maketitle

\begin{abstract}
 Code-switching is a common phenomenon among multilingual speakers, where alternation between two or more languages occurs within the context of a single conversation.
 While multilingual humans can seamlessly switch back and forth between languages, multilingual neural machine translation (NMT) models are not robust to such sudden changes in input. This work explores multilingual NMT models’ ability to handle code-switched text. First, we propose checks to measure switching capability. Second, we investigate simple and effective data augmentation methods that can enhance an NMT model’s ability to support code-switching. Finally, by using a glass-box analysis of attention modules, we demonstrate the effectiveness of these methods in improving robustness.
\end{abstract}

\section{Introduction}

Neural machine translation (NMT)~\cite{sutskever2014sequence,Bahdanau-2015-nmt,vaswani-2017-attention} has made significant progress, from supporting only a pair of languages per model to simultaneously supporting hundreds of languages \cite{johnson-etal-2017-googles,zhang-etal-2020-improving,tiedemann-2020-tatoeba,gowda-etal-2021-many}. 
Multilingual NMT models have been deployed in production systems and are actively used to translate across languages in day-to-day settings \cite{wu2016google,coswell-2020-googletrans,mohan-2021-ms-translator}. 
A great many metrics for evaluation of machine translation have been proposed \cite{doddington-2002-NIST,banerjee-lavie-2005-meteor,snover-etal-2006-study,popovic-2015-chrf,gowda-etal-2021-macro}; simply citing a more comprehensive list would exceed space limitations, 
however, except context-aware MT, nearly all approaches consider translation in the context of a \textit{single sentence}.
Even approaches that generalize to support translation of multiple languages \cite{zhang-etal-2020-improving,tiedemann-2020-tatoeba,gowda-etal-2021-many} continue to use the \textit{single-sentence, single-language} paradigm.
In reality, however, multilingual environments often involve \textit{language alternation} or \textit{code-switching (CS)}, where seamless alternation between two or more languages occurs \cite{cms-and-ury-1977-biling}.

 CS can be broadly classified into two types \cite{myers1989codeswitching}: (i) intra-sentential CS, where switching occurs \textit{within} sentence or clause boundary, and (ii) inter-sentential CS, where switching occurs \textit{at} sentence or clause boundaries.
 An example for each type is given in Table~\ref{tab:cs-example}.
CS has been studied extensively in linguistics communities \cite{nilep-2006-codeswitch}; 
however, the efforts in the MT community are scant \cite{gupta-etal-2021-training}.


\begin{table}[h]
\centering
    \begin{tabular}{c | p{5.6cm}}
     \hline \hline 
    \multirow{2}{*}{\textit{Intra}} & \textsl{\underline{Ce} moment when you start \underline{penser en deux langues} at the same \underline{temps}.} \\ 
    & \small{(The moment when you start to think in two languages at the same time.)} \\ \hline

    \multirow{2}{*}{\textit{Inter}} & \textsl{\underline{Comme on fait son lit}, you must lie on it.} \\
    & \small{(As you make your bed, you must lie on it.)} \\
    \hline \hline 
    \end{tabular}
   
    \caption{Intra- and inter- sentential code-switching examples between \underline{French} and English.}
    \label{tab:cs-example}
\end{table}

In this work, we show that, as commonly built, multilingual NMT models are \textit{not robust} to multi-sentence translation, especially when CS is involved. The contributions of this work are outlined as follows:
Firstly, a few simple but effective checks for improving the test coverage in multilingual NMT evaluation are described (Section~\ref{sec:multiling-mt-checks}).
Secondly, we explore training data augmentation techniques such as concatenation and noise addition in the context of multilingual NMT (Section~\ref{sec:train-aug}).
Third, using a many-to-one multilingual translation task setup (Section~\ref{sec:setup}), we investigate the relationship between training data augmentation methods and their impact on multilingual test cases. 
Fourth, we conduct a glass-box analysis of cross-attention in the Transformer architecture and show visually as well as quantitatively that the models trained with concatenated training sentences learn a more sharply focused attention mechanism than others.
Finally, we examine how our data augmentation strategies generalize to multi-sentence translation for a variable number of sentences, and determine that two-sentence concatenation in training is sufficient to model many-sentence concatenation in inference (Section~\ref{sec:generalization}).

\section{Multilingual Translation Evaluation: Additional Checks}
\label{sec:multiling-mt-checks}


\textit{Notation:} For simplicity, consider a many-to-one model that translates sentences from $K$ source languages,  $\{L_k | k = 1, 2, ... K\}$, to a target language, $T$.
Let $x_{i}^{(L_k)}$ be a sentence in the source language $L_k$, and let its translation in the target language be $y_{i}^{(T)}$; where unambiguous we omit the superscripts.

We propose the following checks to be used for multilingual NMT:
\begin{description}[itemsep=-2mm,topsep=1mm,leftmargin=3mm]
    
    \item[C-TL:] Consecutive sentences in the source and target languages.
    This check tests if the translator can translate in the presence of inter-sentential CS, and preserve phrases that are already in the target language.
    For completeness, we can test both source-to-target and target-to-source CS, as follows: \begin{align}
       x^{(L_k)}_i + y_{i+1} & \rightarrow y_i + y_{i+1} \\
       y_{i} + x^{(L_k)}_{i+1} & \rightarrow y_{i} + y_{i+1}
    \end{align}
    In practice, we use a space character to join sentences, indicated by the concatenation operator `$+$'.\footnote{We focus on orthographies that use space as a word-breaker. In orthographies without a word-breaker, joining may be performed without any glue character.}
  This check requires the held-out set sentence order to preserve the coherency of the original document.
    
    \item[C-XL:] This check tests if a multilingual translator is agnostic to CS. 
    This check is created by concatenating consecutive sentences across source languages. 
    This is possible iff the held-out sets are multi-parallel across languages, and, similar to the previous, each preserves the coherency of the original documents.
    Given two languages $L_k$ and $L_m$, we obtain a test sentence as follows: 
    \begin{equation}
      x^{(L_k)}_i + x^{(L_m)}_{i+1} \rightarrow y_i + y_{i+1}  
    \end{equation}
    
    \item[R-XL:] This check tests if a multilingual translator can function in light of a topic switch among its supported source languages. 
    For any two languages $L_k$ and $L_m$ and random positions $i$ and $j$ in their original corpus, we obtain a test segment by concatenating them as: 
    \begin{equation}
        x^{(L_k)}_i + x^{(L_m)}_j \rightarrow y_i + y_j 
    \end{equation}
     This method makes the fewest assumptions about the nature of held-out datasets, i.e., unlike previous methods, neither multi-parallelism nor coherency in sentence order is necessary.

    \item[C-SL:] Concatenate consecutive sentences in the same language. 
    While this check is not a test on CS, this helps in testing if the model is invariant to a missed segmentation, as it is not always trivial to determine sentence segmentation in continuous language. 
    This check is possible iff held-out set sentence order preserves the coherency of the original document. 
    Formally, 
    \begin{equation}
        x^{(L_k)}_{i} + x^{(L_k)}_{i+1} \rightarrow y_i + y_{i+1}
    \end{equation}
 \end{description}

\section{Achieving Robustness via Data Augmentation Methods}
\label{sec:train-aug}
In the previous section, we described several ways of improving \textit{test} coverage for multilingual translation models.
In this section, we explore \textit{training} data augmentation techniques to improve robustness to code-switching settings.

\subsection{Concatenation}
Concatenation of training sentences has been proven to be a useful data augmentation technique; \citet{nguyen-etal-2021-data} investigate key factors behind the usefulness of training segment concatenations in \textit{bilingual} settings. 
Their experiments reveal that concatenating random sentences performs as well as consecutive sentence concatenation, which suggests that discourse coherence is unlikely the driving factor behind the gains.
They attribute the gains to three factors: context diversity, length diversity, and position shifting.

In this work, we investigate training data concatenation under \textit{multilingual} settings, hypothesizing that concatenation helps achieve the robustness checks that are described in Section~\ref{sec:multiling-mt-checks}. Our training concatenation approaches are similar to our check sets, with the notable exception that we do not consider consecutive sentence training specifically, both because of \citet{nguyen-etal-2021-data}'s finding and because training data gathering techniques can often restrict the availability of consecutive data \cite{banon-etal-2020-paracrawl}.  
We investigate the following sub-settings for concatenations:

\begin{description}[itemsep=0.5mm,topsep=0pt,leftmargin=5mm]
\item[CatSL:] Concatenate a pair of source sentences in the same language, using space whenever appropriate (e.g., languages with space separated tokens). 

\begin{equation}
  x^{(L_k)}_i + x^{(L_k)}_j \rightarrow y_i + y_j  
\end{equation}

\item[CatXL:] Concatenate a pair of source sentences, without constraint on language.
\begin{equation}
  x^{(L_k)}_i + x^{(L_m)}_j \rightarrow y_i + y_j  
\end{equation}
\item[CatRepeat:] The same sentence is repeated and then concatenated. 
Although this seems uninteresting, it serves a key role in ruling out gains possibly due to data repetition and modification of sentence lengths.
\begin{equation}
   x^{(L_k)}_i + x^{(L_k)}_i \rightarrow y_i + y_i 
\end{equation}
\end{description}

\subsection{Adding Noise}
We hypothesize that introducing noise during training might help achieve robustness and investigate two approaches that rely on noise addition:
\begin{description}[itemsep=0.5mm,topsep=0pt,leftmargin=5mm]
\item[DenoiseTgt:] Form the source side of a target segment by adding noise to it. 
Formally, $noise(y; r) \rightarrow y$, where  hyperparameter $r$ controls the noise ratio.
Denoising is an important technique in unsupervised NMT \cite{artetxe2018unsupervised,lample2018unsupervised}.

\item[NoisySrc:] Add noise to the source side of a translation pair.
Formally, $noise(x; r) \rightarrow y$. 
This resembles back-translation~\cite{sennrich-etal-2016-improving} where augmented data is formed by pairing noisy source sentences with clean target sentences.

\end{description}

The function $noise(...; r)$ is implemented as follows: 
(i) $r\%$ of random tokens are dropped, 
(ii) $r\%$ of random tokens are replaced with random types uniformly sampled from vocabulary, and 
(iii) $r\%$ of random tokens' positions are displaced within a sequence.
We use $r=10\%$ in this work.

\renewcommand{\arraystretch}{1.2}
\begin{table}[htb]
    \centering
\small
\begin{tabular}{l @{\hspace{6pt}} c @{\hspace{6pt}} r }
\hline
\textbf{Language} & \textbf{In-domain} & \multicolumn{1}{c}{\textbf{All-data}} \\
\hline
Bengali (BN)  & 23.3k/0.4M/0.4M    & 1.3M/19.5M/21.3M \\
Gujarati (GU) & 41.6k/0.7M/0.8M    & 0.5M/07.2M/09.5M \\
Hindi (HI)    & 50.3k/1.1M/1.0M    & 3.1M/54.7M/51.8M \\
Kannada (KN)  & 28.9k/0.4M/0.6M    & 0.4M/04.6M/08.7M \\
{Malayalam}(ML) & 26.9k/0.3M/0.5M  & 1.1M/11.6M/19.0M \\
Marathi (MR)  & 29.0k/0.4M/0.5M    & 0.6M/09.2M/13.1M \\
Oriya (OR)    & 32.0k/0.5M/0.6M    & 0.3M/04.4M/05.1M \\
Punjabi (PA)  & 28.3k/0.6M/0.5M    & 0.5M/10.1M/10.9M \\
Tamil (TA)    & 32.6k/0.4M/0.6M    & 1.4M/16.0M/27.0M \\
Telugu (TE)   & 33.4k/0.5M/0.6M    & 0.5M/05.7M/09.1M \\
\hdashline
All   & 326k/5.3M/6.1M     & 9.6M/143M/175M\\ 
\hline
\end{tabular} 
    \caption{Training dataset statistics: \textit{segments / source / target tokens}, before tokenization.}
    \label{tab:training-stats}
    \centering
    \small
\begin{tabular}{l c c}
\hline
\textbf{Name} & \textbf{Dev} & \textbf{Test} \\
\hline
Orig & 10k/140.5k/163.2k & 23.9k/331.1k/385.1k \\
C-TL & 10k/303.7k/326.4k & 23.9k/716.1k/770.1k \\
C-XL & 10k/283.9k/326.4k & 23.9k/670.7k/770.1k \\
R-XL & 10k/216.0k/251.2k & 23.9k/514.5k/600.5k \\
C-SL & 10k/281.0k/326.4k & 23.9k/662.1k/770.1k \\
\hline
\end{tabular}
    \caption{Development and test set statistics: \textit{segments / source / target tokens}, before subword tokenization. 
    The row named `Orig' is the union of all ten individual languages' datasets, and the rest are created as per definitions in Section~\ref{sec:multiling-mt-checks}.
    Dev-Orig set is used for validation and early stopping in all our multilingual models.}
    \label{tab:heldout-stats}
\end{table}

\begin{table}[htb]
    \centering
    \includegraphics[width=\linewidth,trim={10mm 2mm 10mm 2mm},clip]{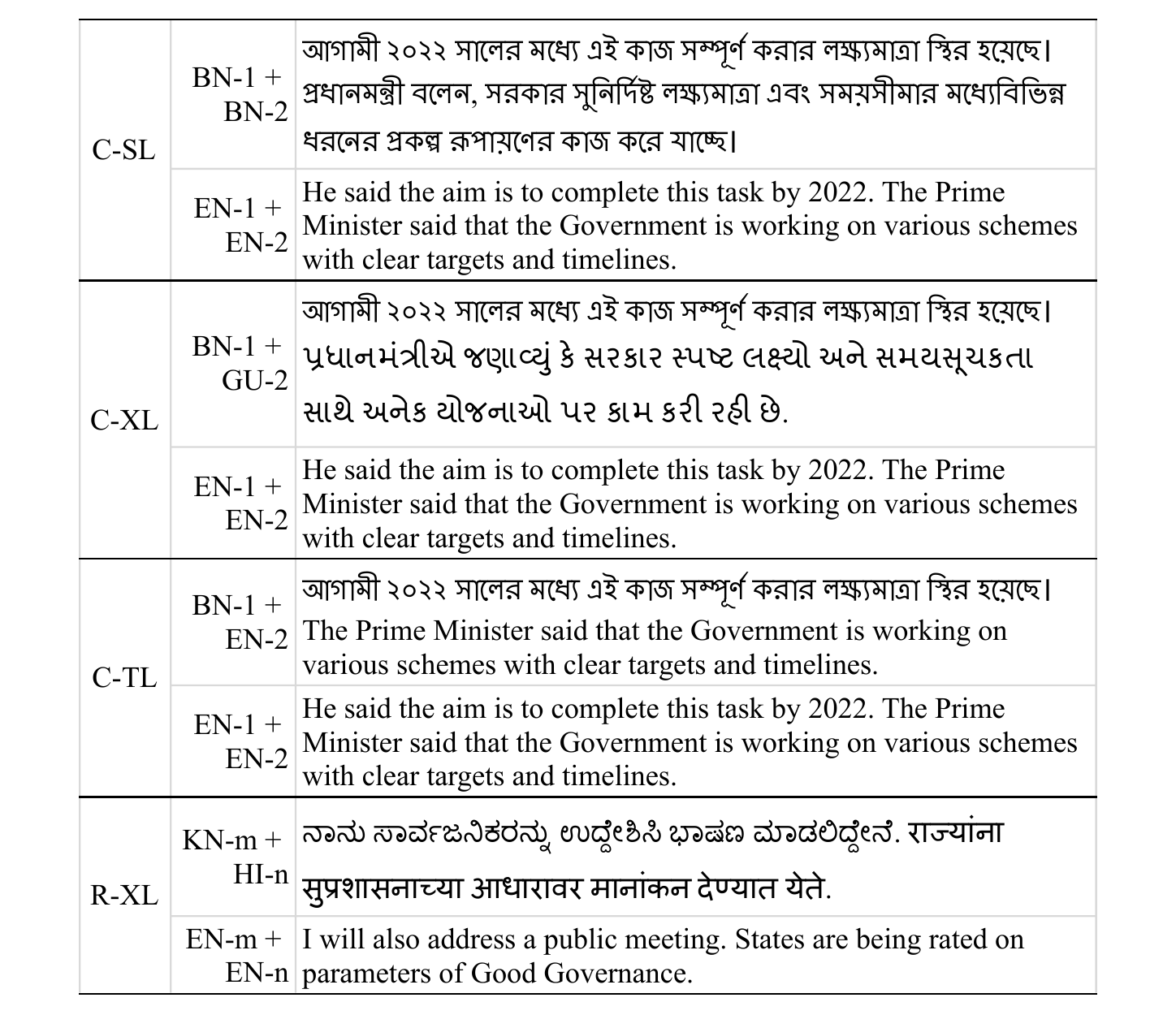}
    \caption{Concatenated sentence examples from the development set. 
    Bengali (BN), Gujarati (GU), Kannada (KN), and Hindi (HI) are chosen for illustrations; similar augmentations are performed for all other languages in the corpus.
    Indices $1$ and $2$ indicate consecutive positions, and $m$ and $n$ indicate random positions.}
    \label{tab:dev-aug-example}
\end{table}

\section{Setup}
\label{sec:setup}

\subsection{Dataset}
We use publicly available datasets from The Workshop on Asian Translation 2021 (WAT21)'s  \textit{MultiIndicMT}~\cite{nakazawa-etal-2021-overview}\footnote{\url{http://lotus.kuee.kyoto-u.ac.jp/WAT/indic-multilingual/}} shared task. This task involves translation between English(EN) and 10 Indic Languages, namely: Bengali(BN), Gujarati(GU), Hindi(HI), Kannada(KN), Malayalam(ML), Marathi(MR), Oriya(OR), Punjabi(PA), Tamil(TA) and Telugu(TE). 
The development and held-out test sets are multi-parallel and contain 1,000 and 2,390 sentences, respectively. 
The training set contains a small portion of data from the same domain as the held-out sets, as well as additional datasets from other domains.
All the training data statistics are given in Table~\ref{tab:training-stats}.
We focus on the Indic$\shortarrow$English (many-to-one) translation direction in this work.

Following the definitions in Section~\ref{sec:multiling-mt-checks}, we create C-SL, C-TL, C-XL, and R-XL versions of development and test sets; statistics are given in Table~\ref{tab:heldout-stats}. 
An example demonstrating the nuances in all these four methods is shown in Table~\ref{tab:dev-aug-example}. 
Following the definitions in Section~\ref{sec:train-aug}, we create CatSL, CatXL, CatRpeat, DenoiseTgt, and NoisySrc augmented training segments. 
For each of these training corpus augmentation methods, we restrict the total augmented sentences to be roughly the same number of segments as the original corpus, i.e., 326k and 9.6M segments in the in-domain and the all-data setup, respectively.

\subsection{Model and Training Process}
We use a Transformer base model \cite{vaswani-2017-attention} which has 512 hidden dimensions, 6 encoder and decoder layers, 8 attention heads, and intermediate feedforward layers of 2048 dimensions.
We use a Pytorch based NMT toolkit.\footnote{Additional details are withheld at the moment to preserve the anonymity of authors. All code, data, and models will be publicly released.} 
 Tuning the vocabulary size and batch size are important to achieve competitive performance.
We use byte-pair-encoding (BPE) \cite{sennrich-etal-2016-neural}, with vocabulary size adjusted as per the recommendations from \citet{gowda-may-2020-finding}. 
Since the source side has many languages and the target side has only a single language, we use a larger source vocabulary than that of the target.
The source side vocabulary contains BPE types from all 11 languages (i.e., ten source languages and English), whereas to improve the efficiency in the decoder's softmax layer, the target vocabulary is restricted to contain English only. 
Our in-domain limited-data setup learns BPE vocabularies of 30.4k and 4.8k types for source and target languages.
Similarly, the all-data setup learns 230.4k and 63.4k types.
The training batch size used for all our multilingual models is 10k tokens for the in-domain limited-data setup, and 25k tokens for the larger all-data setup.
The batch size for the baseline bilingual models is adjusted as per data sizes using \textit{`a thousand per million tokens'} rule of thumb that we have come to devise with a maximum of 25k tokens.
 The median sequence lengths in training after subword segmentation but before sentence concatenation are 15 on the Indic side and 17 on the English side. 
 We model sequence lengths up to 512 time steps during training.
We use the same learning rate schedule as \citet{vaswani-2017-attention}.
We train our models until a maximum of 200k optimizer steps, and use early stopping with a patience of 10 validations.
Validations are performed after every 1000 optimizer steps.
All our models are trained using one Nvidia A40 GPU per setting. 
The smaller in-domain setup takes less than 24 hours per run, whereas the larger all-data setup takes at most 48 hours per run (or less when early stopping criteria are reached).
We run each experiment two times and report the average. During inference, we average the last 5 checkpoints and use a beam decoder of size 4 and length penalty of $\alpha=0.6~$\cite{vaswani-2017-attention,wu2016google}.

\begin{table*}[h!t]
\centering
\small
\centering
\small
\begin{tabular}{l l rrrrr : rrrrr}
\hline
& & \multicolumn{5}{c:}{\textbf{Dev}} & \multicolumn{5}{c}{\textbf{Test}} \\
ID & In-domain & \multicolumn{1}{r}{Orig} & C-TL & C-SL & C-XL & R-XL & Orig & C-TL & C-SL & C-XL & R-XL \\
\hline
\#I1 & Baseline (B) & 26.5 & 10.8 & 17.0 & 16.9 & 15.9  & 22.7 & 9.4 & 14.9 & 14.7 & 13.6 \\
\hdashline
\#I2 & B+CatRepeat & 25.3 & 9.9 & 14.5 & 14.7 & 13.3 & 21.6 & 8.6 & 13 & 13 & 11.4 \\ 
\#I3 & B+CatXL     & 26.2 & 12.6 & 26.1 & 25.9 & \textbf{26.5} & 22.6 & 11.1 & 22.7 & 22.5 & 22.3 \\
\#I4 & B+CatSL     & 26.1 & 13.2 & 26.1 & 25.9 & \textbf{26.5} & 22.6 & 11.4 & 22.9 & \textbf{22.6} & 22.3 \\
\#I5 & B+NoisySrc  & 25.2 & 10.5 & 16.2 & 16.0 & 15.2 & 21.2 & 9.1 & 14.3 & 14.1 & 12.9 \\
\#I6 & B+DenoiseTgt & \textbf{26.7} & 40.4 & 17.9 & 17.7 & 16.6 & \textbf{23.2} & 39.7 & 15.7 & 15.4 & 14.1 \\
\hdashline
\#I7 & B+CatXL+DenoiseTgt & 26.1 & \textbf{55.2} & \textbf{26.3} & \textbf{26.0} & 26.4 & 22.6 & \textbf{53.4} & \textbf{23.0} & \textbf{22.6} & \textbf{22.4} \\ 
\hline
\end{tabular} 
\caption{ Indic$\shortarrow$English BLEU scores for models trained on in-domain training data only. \textit{Abbreviations:} Orig: average across ten languages' original held-out set, C-: consecutive sentences, R-: random sentences, TL: target-language (i.e, English), SL: same-language, XL: cross-language. }
\label{tab:bleu-indom-aug}
\vspace{4mm}
\begin{tabular}{l l rrrrr : rrrrr}
\hline
& & \multicolumn{5}{c:}{\textbf{Dev}} & \multicolumn{5}{c}{\textbf{Test}} \\
ID & All-data & Orig & C-TL & C-SL & C-XL & R-XL & Orig & C-TL & C-SL & C-XL & R-XL \\
\hline
\#A1 & Baseline (B) & \textbf{35.0} & 43.1 & 30.0 & 29.5 & 28.2 & \textbf{32.4} & 42.2 & 27.8 & 27.3 & 26.1 \\
\hdashline
\#A2 & B+CatRepeat & 34.5 & 43.7 & 30.3 & 29.9 & 28.8 & 32.0 & 42.9 & 28.0 & 27.6 & 26.3 \\ 
\#A3 & B+CatXL & 34.1 & 53.3 & 31.9 & \textbf{33.7} & \textbf{34.4} & 31.6 & 52.4 & 29.7 & \textbf{31.0} & \textbf{31.2} \\
\#A4 & B+CatSL & 33.6 & 54.0 & \textbf{32.5} & 32.2 & 34.3 & 31.3 & 53.3 & \textbf{30.4} & 29.9 & 31.1 \\
\#A5 & B+NoisySrc & 34.9 & 42.1 & 29.8 & 29.2  & 27.8 & 32.3 & 41.7 & 27.6 & 27.1 & 25.8 \\
\#A6 & B+DenoiseTgt & 33.3 & 60.0 & 28.9 & 28.4 & 27.3 & 31.3 & 59.4 & 27.1 & 26.5 & 25.4 \\
\hdashline
\#A7 & B+CatXL+DenoiseTgt & 33.3 & \textbf{65.8} & 31.1 & 33.0 & 33.6 & 31.0 & \textbf{64.7} & 28.9 & 30.4 & 30.3 \\ 
\hline
\end{tabular} 
\caption{Indic$\shortarrow$English BLEU scores for models trained on all data. (Abbreviations are same as Table~\ref{tab:bleu-indom-aug}.) }
\label{tab:bleu-alldata-augs}
\vspace{4mm}
\centering
\small
\begin{tabular}{l l : rrrr : rrrr}\hline
     &  & \multicolumn{4}{c:}{\textbf{Dev}} & \multicolumn{4}{c}{\textbf{Test}} \\
          ID &      & C-TL & C-SL & C-XL & R-XL & C-TL & C-SL & C-XL & R-XL \\ \hline
\#A1 & Baseline (B) & 14.3 & 10.4 & 10.3 & 10.1 & 14.3 & 10.6 & 10.5 & 10.3 \\ \hdashline
\#A2 & B+CatRepeat  & 12.3 & 8.9 & 8.9 & 8.6 & 12.5 & 9.0 & 9.0 & 8.7 \\
\#A3 & B+CatXL      &  {5.8} & 7.2 & 4.3 & 4.3 & 5.8 & {7.2} & {4.4} & {4.3} \\
\#A4 & B+CatSL      & 5.3 & \textbf{6.2} & 6.1 & 5.2 & {5.4} & \textbf{6.2} & 6.2 & 5.2 \\
\#A5 & B+NoisySrc   & 17.4 & 16.1 & 16.1 & 15.8 & 17.5 & 16.2 & 16.2 & 15.9 \\
\#A6 & B+DenoiseTgt & 7.9 & 8.3 & 8.4 & 8.0 & 8.1 & 8.5 & 8.5 & 8.1 \\
\hdashline
 \#A7 & B+CatXL+DenoiseTgt & \textbf{4.3} & {6.8} & \textbf{3.9} & \textbf{4.1} & \textbf{4.4} & {7.0} & \textbf{4.0} & \textbf{4.1} \\ 
\hline
\end{tabular} 
\caption{Cross-attention bleed rate (lower is better). All numbers are scaled from $[0,1]$ to $[0, 100]$ for easier interpretation, and the best settings per test are indicated with bold font.
Models trained on concatenated sentences have lower attention bleed rate. Denoising is better than baseline, but not as much as concatenation. 
The lowest bleed rate is achieved by using both concatenation and denoising methods. (Abbreviations are same as Table~\ref{tab:bleu-indom-aug}.)}
\label{tab:xattn-bleed}
\end{table*}

\section{Results and Analysis}
\label{sec:results-and-analysis}

We train multilingual many-to-one models for both in-domain and all data. Table~\ref{tab:bleu-indom-aug} presents our results from a limited quantity in-domain dataset.
The baseline model (\#I1) has strong performance on individual sentences, but degrades on held-out sets involving missed sentence segmentation and code-switching.
Experiments with concatenated data, namely CatXL (\#I3) and CatSL (\#I4), while they appear to make no improvements on regular held-out sets, make a significant improvement in BLEU scores on C-SL, C-XL, and R-XL.
Furthermore, both CatSL and CatXL show a similar trend.
While they also make a small gain on the C-TL setting, DenoiseTgt method is clearly an out-performer on C-TL.
The model that includes both concatenation and denoising (\#I7) achieves consistent gains across all the robustness check columns.
In contrast, the CatRepeat (\#I2) and NoisySrc (\#I5) methods do not show any gains.

Our results from the all-data setup are provided in Table~\ref{tab:bleu-alldata-augs}. 
While none of the augmentation methods appear to surpass baseline BLEU on the regular held-out sets (i.e., Orig column), their improvements to robustness can be witnessed similar to the in-domain setup. 
We show a qualitative example in Table~\ref{tab:fig:example-translations}.

\begin{table}[htb]
    \centering
    \includegraphics[width=\linewidth,trim={4mm 3mm 4mm 4mm},clip]{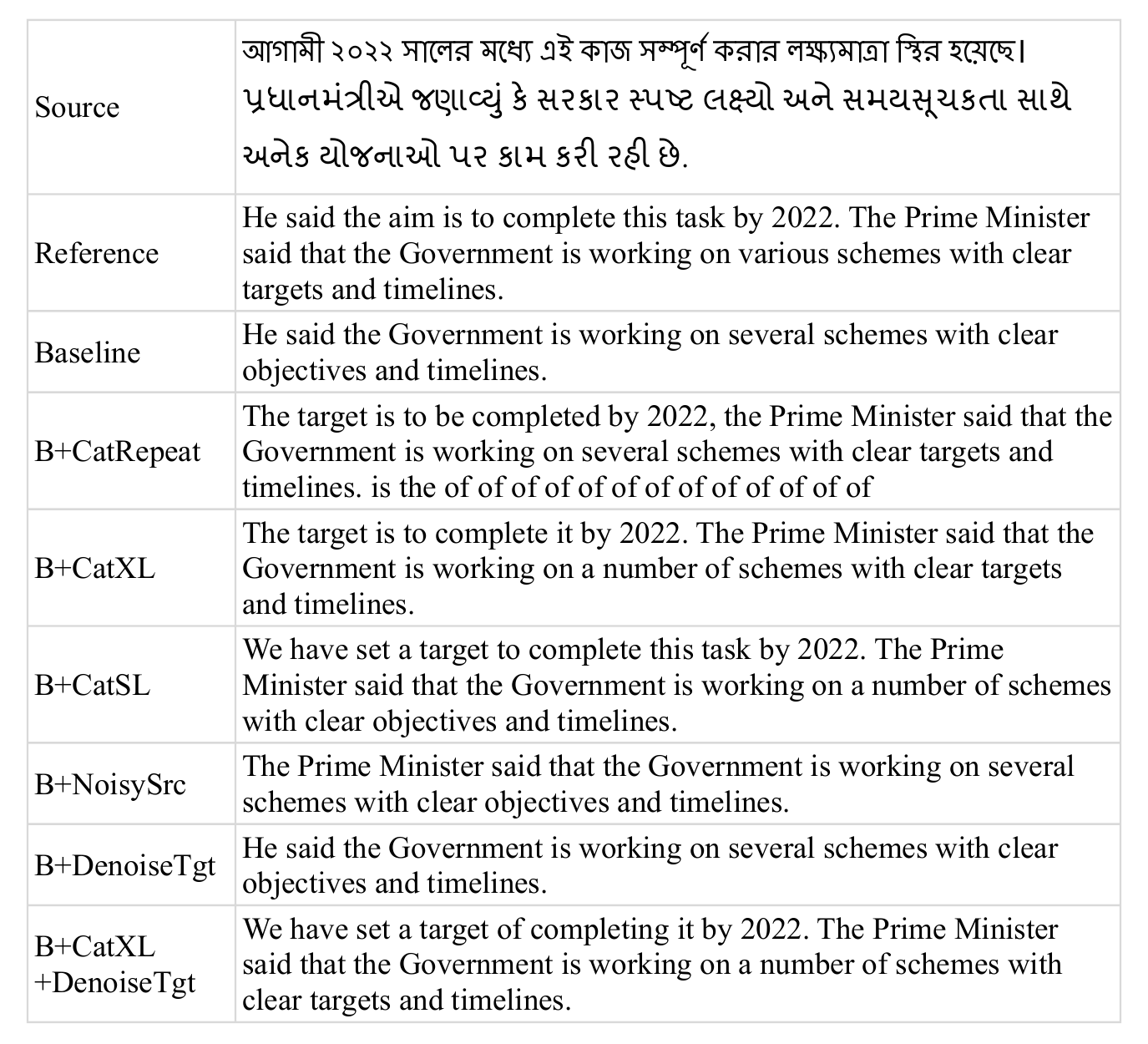}
    \caption{Example translations from the models trained on all-data setup.
    See Table~\ref{tab:bleu-alldata-augs} for quantitative scores of these models, and Figures \ref{fig:attnviz-catXL} and \ref{fig:attnviz-catXL+denoise} for a visualization of cross-attention.}
    \label{tab:fig:example-translations}
\end{table}

\subsection{Attention Bleed}
\label{sec:attn-bleed}

\begin{figure}[htb!]
    \centering
    \begin{subfigure}[b]{0.99\linewidth}
        \centering
        \includegraphics[width=\linewidth,trim={0mm 3mm 38mm 3mm},clip]{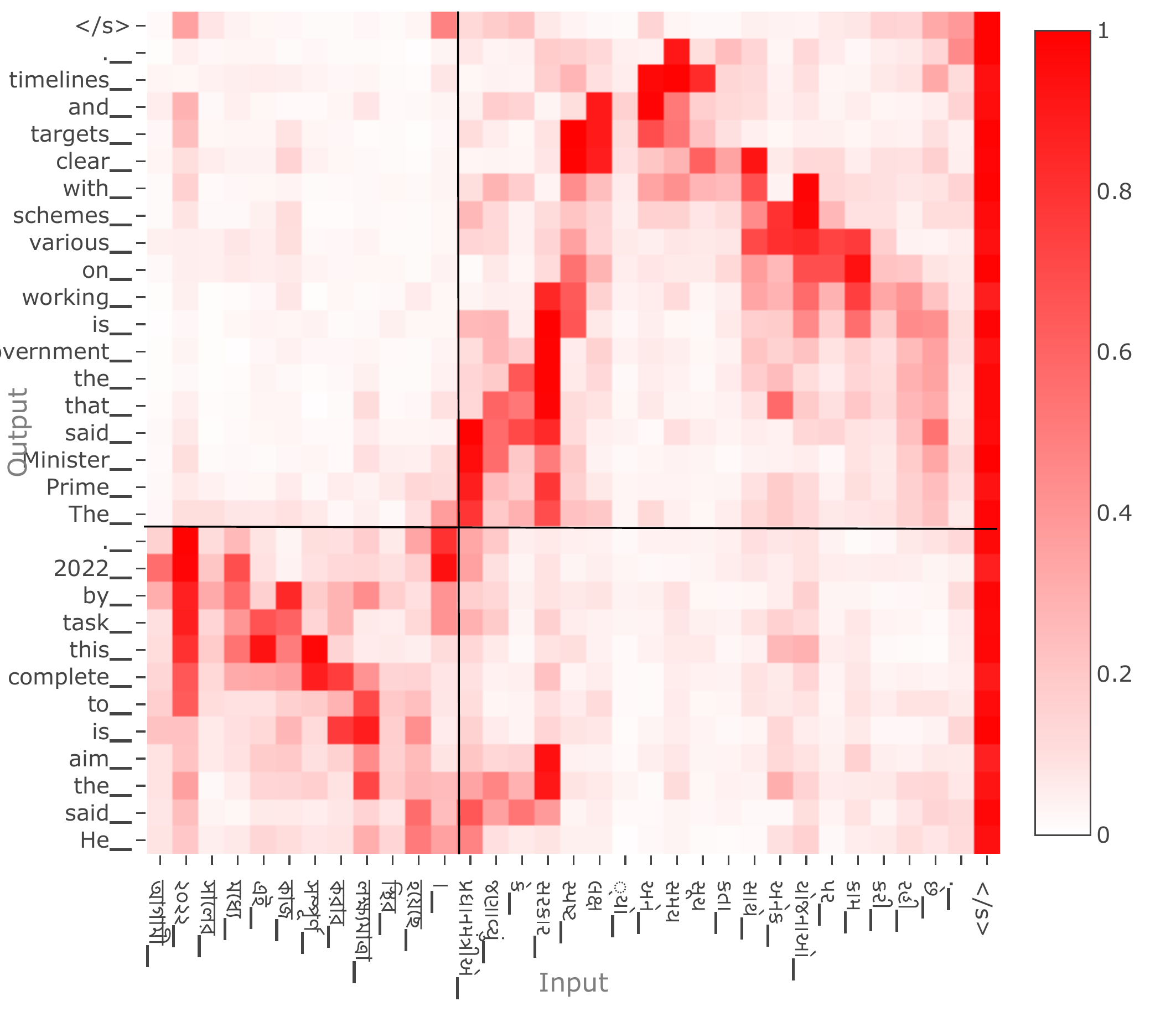}
        \caption{Baseline model without sentence concatenation (\#A1)}
        \label{fig:baseline-on-catXL}
     \end{subfigure}
    
    \vspace{5mm}
    
    \begin{subfigure}[b]{0.99\linewidth}
        \centering
        \includegraphics[width=\linewidth,trim={0mm 3mm 38mm 1mm},clip]{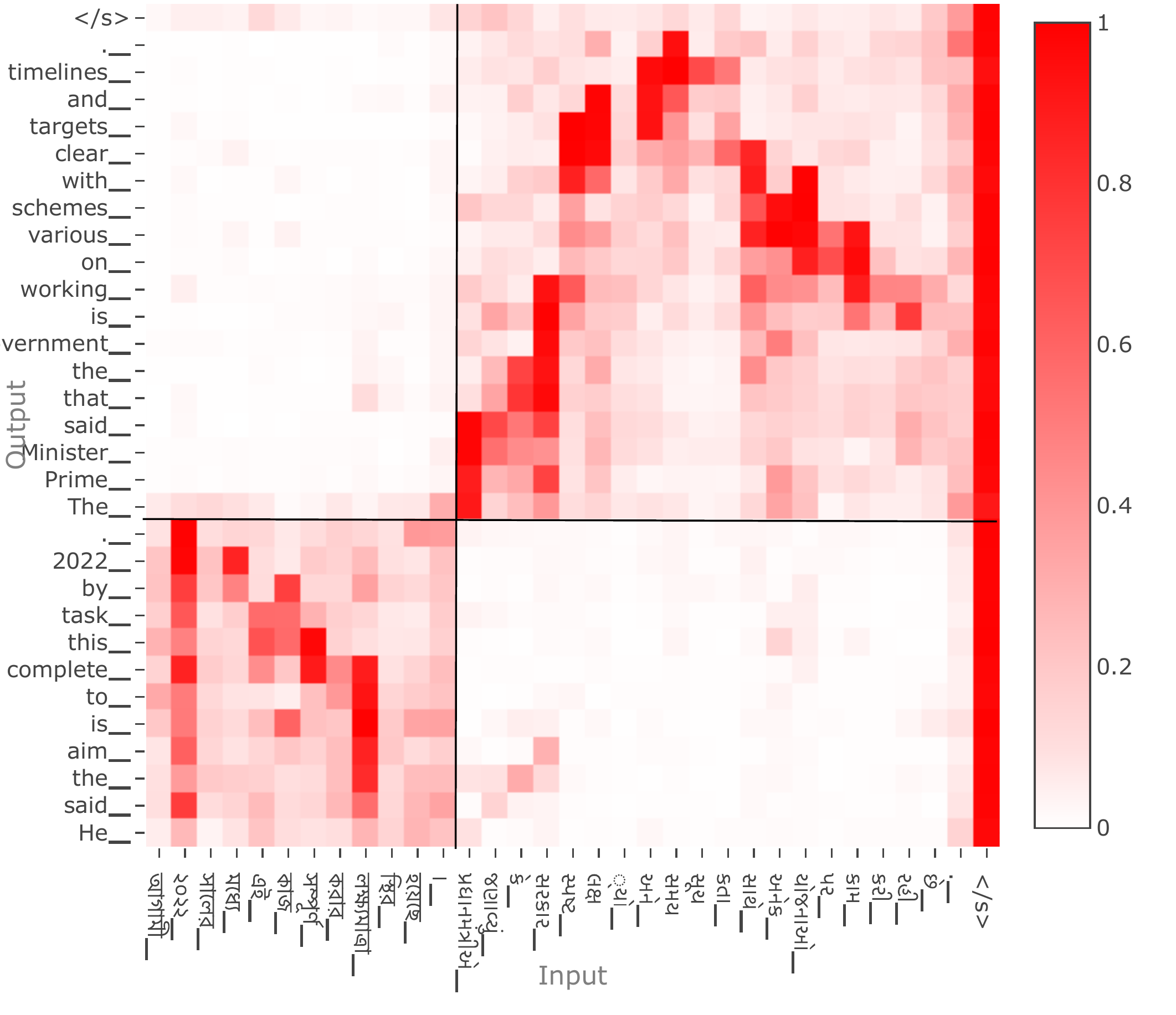}
        \caption{Model trained with concatenated sentences (\#A3)}
        \label{fig:CatXL-on-CatXL}
     \end{subfigure}
   
    \caption{Cross-attention visualization from baseline model and concatenated (cross-language) model.
     For each position in the grid, only the maximum value across all attention-heads from all the layers is visualized. The darker color implies more attention weight, and the black bars indicate sentence boundaries.
    The model trained on concatenated sentences has more pronounced cross-attention boundaries than the baseline, indicating less mass is bled across sentences.}
\label{fig:attnviz-catXL}
\end{figure}

\begin{figure}[h!t]
    \centering
    \begin{subfigure}[b]{\linewidth}
         \centering
         \includegraphics[width=\linewidth,trim={0mm 3mm 38mm 2mm},clip]{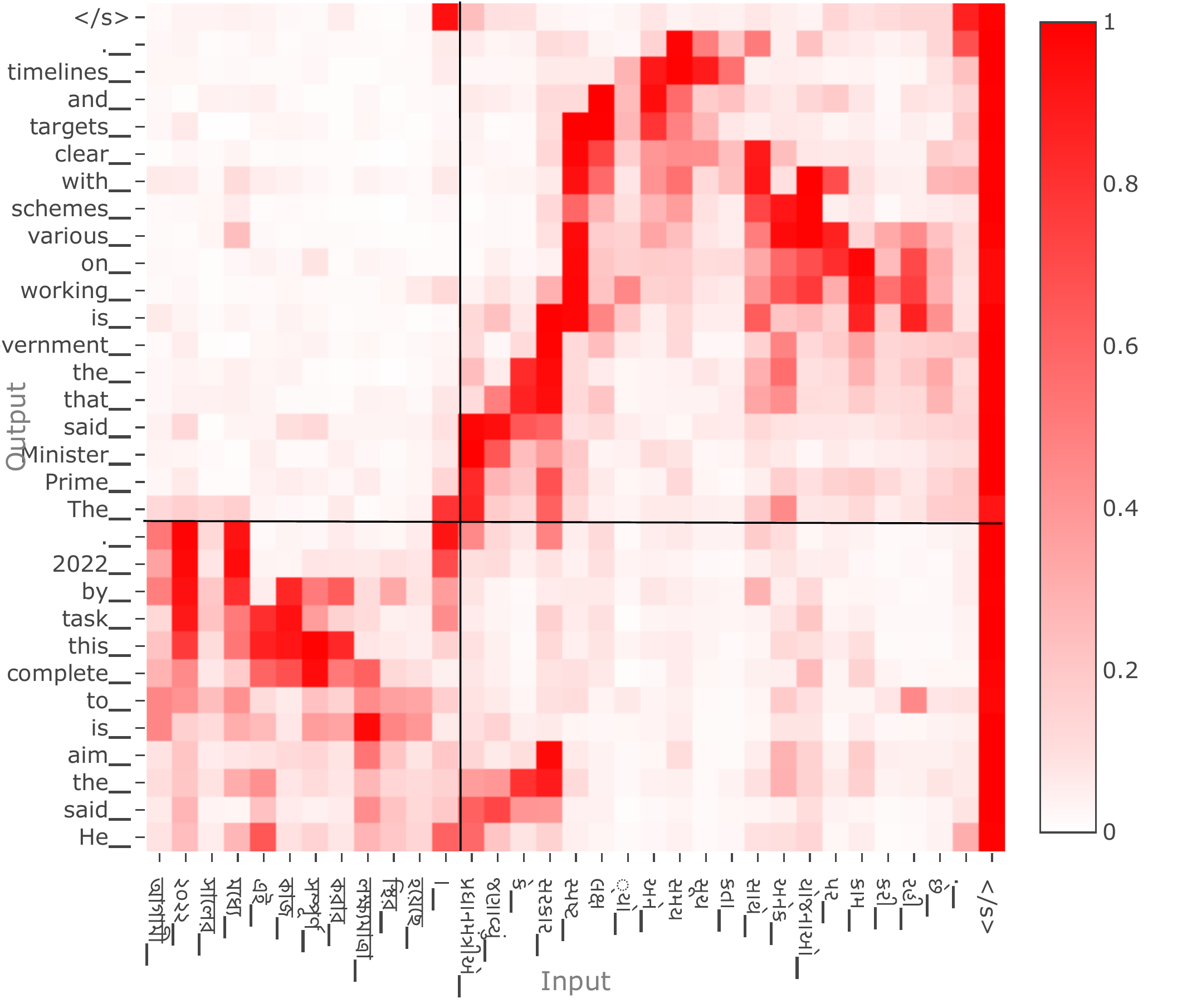}
         \caption{Model trained with DenoiseTgt augmentation (\#A6)}
         \label{fig:denoise-on-catXL}
      \end{subfigure}
      
      \vspace{5mm}
      
     \begin{subfigure}[b]{\linewidth}
         \centering
         \includegraphics[width=\linewidth,trim={0mm 3mm 38mm 2mm},clip]{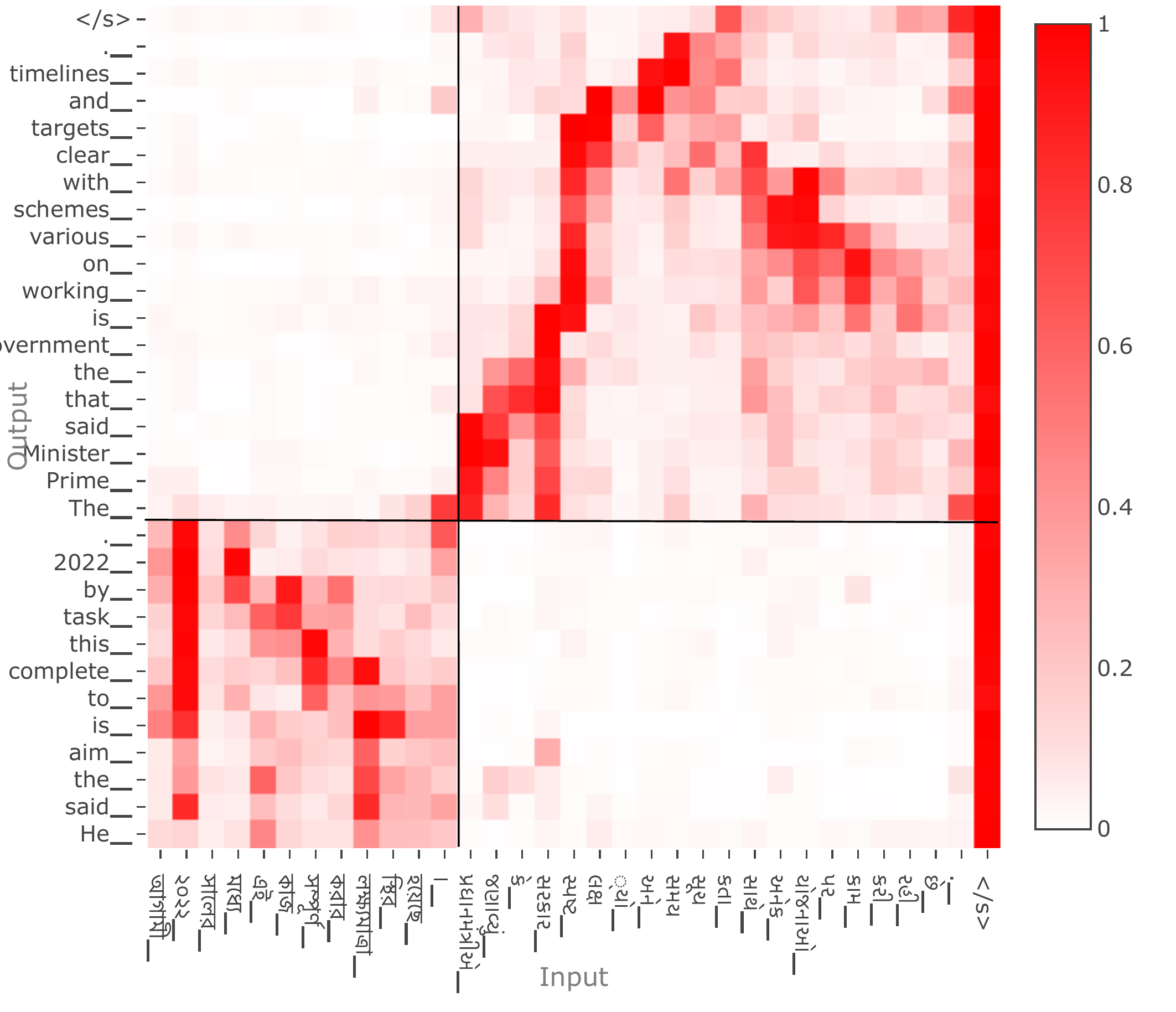}
         \caption{Model trained with both CatXL and DenoiseTgt augmentations (\#A7)}
         \label{fig:denoise+catXL-on-catXL}
      \end{subfigure}
    \caption{Cross-attention visualization (... continuation from Figure~\ref{fig:attnviz-catXL}) The model trained on both concatenated and denoising sentences has least attention mass across sentences.}
    \label{fig:attnviz-catXL+denoise}
\end{figure}

Figures~\ref{fig:attnviz-catXL} and \ref{fig:attnviz-catXL+denoise} visualize cross-attention\footnote{Also known as encoder-decoder attention.} from our baseline model without augmentation as well as models trained with augmentation. 
Generally, the NMT decoder is run autoregressively; however, to facilitate the analysis described in this section, we force-decode reference translations and extract cross-attention tensors from all models.
The cross-attention visualization between a pair of concatenated sentences, say $(x_{i1} + x_{i2} \rightarrow y_{i1} + y_{i2})$, shows that models trained on augmented datasets appear to have less cross-attention mass across sentences, i.e. in the attention grid regions representing $x_{i2} \leftarrow y_{i1}$, and $x_{i1} \leftarrow y_{i2}$. 
We call attention mass in such regions \textit{attention bleed}. 
This observation confirms some of the findings suggested by \citet{nguyen-etal-2021-data}. 
We quantify attention bleed as follows: 
consider a Transformer NMT model with $L$ layers, each having $H$ attention heads and a held-out dataset of $\{(x_i ~ y_i) | i=1,2,...N\}$ segments. 
Further more, let each segment $(x_i, y_i) $ be a concatenation of two sentences i.e. $(x_{i1}+ x_{i2}, ~ y_{i1}+y_{i2})$, with known sentence boundaries.
Let $|x_i|$ and $|y_i|$ be the sequence lengths after BPE segmentation, and $|x_{i1}|$ and $|y_{i1}|$ be the indices of the end of the first sentence (i.e., the sentence boundary) on the source and target sides, respectively.
The average attention bleed across all the segments, layers, and heads is defined as:
\begin{align}
    \bar{B} = \frac{1}{N \times L \times H} \sum_{i=1}^N \sum_{l=1}^L \sum_{h=1}^H b_{i,l,h}
\end{align}

where $b_{i,l,h}$ is the attention bleed rate in an attention head $h\in[1, H]$, in layer $l \in [1, L]$, for a single record at $i\in[1,N]$. 
To compute $b_{i,l,h}$, consider that an attention grid $A^{(i,l,h)}$ is of size $|y_i|\times|x_i|$. Then 
\begin{multline}
b_{i,l,h} = \frac{1}{|y_i|} \Big[
      \sum_{t=1}^{|y_{i1}|} \sum_{s=|x_{i1}|+1}^{|x_i|} A^{(i,l,h)}_{t,s}  + \\
      \sum_{t=|y_{i1}|+1}^{|y_i|} \sum_{s=1}^{|x_{i1}|} A^{(i,l,h)}_{t,s} \Big]  
\end{multline} 


\noindent where $A^{(i,l,h)}_{t,s}$ is the percent of attention paid to source position $s$ by target position $t$ at decoder layer $l$ and head $h$ in record $i$. Intuitively, a lower value of $\bar{B}$ is better, as it indicates that the model has learned to pay attention to appropriate regions. 
 As shown in Table~\ref{tab:xattn-bleed}, the models trained on augmented sentences achieve lower attention bleed. 

\subsection{Sentence Concatenation Generalization}
\label{sec:generalization}
In the previous sections, only two-segment concatenation has been explored; here, we investigate whether more concatenation further improves model performance and whether models trained on two segments generalize to more than two at test time.
We prepare a training dataset having up to four sentence concatenations and evaluate on datasets having up to four sentences.
As shown in Table~\ref{tab:cat-gen-bleus}, the model trained with just two segment concatenation achieves a similar BLEU as model trained with up to four concatenations.
\begin{table}[ht]
\centering
\small
\begin{tabular}{l rr : rr}
\hline
&\multicolumn{2}{c}{\textbf{Dev}} & \multicolumn{2}{c}{\textbf{Test}} \\
& C-SL & C-4SL & C-SL & C-4SL \\
\hline
Baseline / no join & 30.0 & 27.8 & 27.8 & 25.7 \\
Up to two joins & 31.9 & 28.9 & 29.7 & 26.7\\
Up to four joins & 31.0 & 28.9 & 28.8 & 26.8\\
\hline
\end{tabular} 
\caption{Indic$\shortarrow$English BLEU on held out sets containing up to 4 consecutive sentence concatenations in same language (C-4SL).
The two sentences dataset (C-SL) is also given for comparison.
The model trained on two concatenated sentences achieves comparable results on C-4SL, indicating that no further gains are obtained from increasing concatenation in training.}
\label{tab:cat-gen-bleus}
\end{table}

\section{Related Work}

\paragraph{Robustness and Code-Switching:}

MT robustness has been investigated before within the scope of bilingual translation settings.
Some of those efforts include robustness against input perturbations \cite{cheng-etal-2018-towards}, naturally occurring noise \cite{vaibhav-etal-2019-improving}, and domain shift \cite{muller-etal-2020-domain}.
However, as we have shown in this work, multilingual translation models can introduce new aspects of robustness to be desired and evaluated.
The robustness checklist proposed by \citet{ribeiro-etal-2020-beyond} for NLP modeling in general does not cover translation tasks, whereas our work focuses entirely on the multilingual translation task. 
\citet{clinchant-etal-2019-use} and \citet{niu-etal-2020-evaluating} create synthetic test sets to increase test coverage, however, unlike our work, their synthetic tests do not simulate CS.
\citet{belinkov2018synthetic} investigate the effect of noise on character based NMT and find that excess noise is detrimental to performance as models are brittle.
\citet{yang-etal-2020-csp} artificially create CS text via unsupervised lexicon induction for pretraining NMT in bilingual settings, and \citet{song-etal-2019-code} make CS training data to achieve lexically constrained translation, however, neither of these investigate model's ability to translate CS text at evaluation time. 

\paragraph{Augmentation Through Concatenation:} Concatenation has been used before as a simple-to-incorporate augmentation method. Concatenation can be limited to consecutive sentences as a means to provide extended context for translation \cite{tiedemann-scherrer-2017-neural,Agrawal2018ContextualHI}, or additionally include putting random sentences together, which has been shown to result in gains under low resource settings \cite{nguyen-etal-2021-data,kondo-etal-2021-sentence}. While in a multilingual setting such as ours, data scarcity is less of a concern as a result of combining multiple corpora, concatenation is still helpful to prepare the model for scenarios where code-switching is plausible.
Besides data augmentation, concatenation has also been used to train multi-source NMT models. Multi-source models \cite{och-ney-2001-statistical} translate multiple semantically-equivalent source sentences into a \textit{single} target sentence. \newcite{DBLP:journals/corr/DabreCK17} show that by concatenating the source sentences (equivalent sentences from different languages), they are able to train a single-encoder NMT model that is competitive with models that use separate encoders for different source languages.
Backtranslation \cite{sennrich-etal-2016-improving} is another useful method for data augmentation, however it is more expensive when the source side has many languages, and does not focus on code-switching. 

\paragraph{Attention Weights:} Attention mechanism \cite{Bahdanau-2015-nmt} enables the NMT decoder to choose which part of the input to \textit{focus} on during its stepped generation. The attention distributions learned while training a machine translation model, as an indicator of the context on which the decoder is focusing, have been used to obtain word alignments \cite{garg-etal-2019-jointly,DBLP:journals/corr/abs-1901-11359,zenkel-etal-2020-end,chen-etal-2020-accurate}. In this work, by visualizing attention weights, we depict how augmenting the training data guides attention to more neatly focus on the sentence of interest while decoding its corresponding target sentence. We are also able to quantify this by the introduction of the attention bleed metric.

\section{Conclusion}

We have described simple but effective checks for improving test coverage in multilingual NMT (Section~\ref{sec:multiling-mt-checks}), and have explored training data augmentation methods such as sentence concatenation and noise addition (Section~\ref{sec:train-aug}).
Using a many-to-one multilingual setup, we have investigated the relationship between these augmentation methods and their impact on robustness in multilingual translation. 
While the methods are useful in limited training data settings, their impact may not be visible on single-sentence test sets in a high resource setting. 
However, our proposed evaluation checks reveals the robustness improvement in both the low resource as well as high resource settings.
We have conducted a glass-box analysis of cross-attention in Transformer NMT showing both visually and quantitatively that the models trained with augmentations, specifically, sentence concatenation and target sentence denoising, learn a more sharply focused attention mechanism (Section~\ref{sec:attn-bleed}).
Finally, we have determined that two-sentence concatenation in training corpora generalizes sufficiently to many-sentence concatenation inference (Section~\ref{sec:generalization}).  

\section*{Acknowledgements}
This research is based upon work supported by the Office of the Director of National Intelligence (ODNI), Intelligence Advanced Research Projects Activity (IARPA), via AFRL Contract FA8650-17-C-9116.  The views and conclusions contained herein are those of the authors and should not be interpreted as necessarily representing the official policies or endorsements, either expressed or implied, of the ODNI, IARPA, or the U.S. Government. The U.S. Government is authorized to reproduce and distribute reprints for Governmental purposes notwithstanding any copyright annotation thereon.

\section*{Limitations}

\begin{enumerate}[itemsep=-1mm,topsep=0mm]
   \item This work is focused on translating CS input, and does not attempt to generate CS text during translation. We consider the general problem of many-to-many translation with CS text on both input and output as a promising future direction.
   
    \item As mentioned in Section~\ref{sec:multiling-mt-checks}, some of the multilingual evaluation checks require the datasets to have multi-parallelism, and coherency in the sentence order. When neither multi-parallelism nor coherency in the held-out set sentence order is available, we recommend R-XL. 
    \item While the proposed checks serve as starting points for testing CS, we do not claim that they are exhaustive of all manner of CS. The proposed checks specifically simulate  inter-sentential CS; intra-sentential CS checks are left for  future work.
    
    \item We have investigated robustness under Indic-English translation tasks where all languages use space characters as word-breakers; we have not investigated other languages such as Chinese, Thai, etc. We use the term \textit{Indic} language to collectively reference 10 Indian languages only, similar to \textit{MultiIndicMT} shared task. While the remaining Indian languages and their dialects are not covered, we believe that the approaches discussed in this work generalize to other languages in the same family. 
\end{enumerate}




\bibliographystyle{acl_natbib}
\bibliography{refs} 

\begin{thebibliography}{45}
\expandafter\ifx\csname natexlab\endcsname\relax\def\natexlab#1{#1}\fi

\bibitem[{Agrawal et~al.(2018)Agrawal, Turchi, and
  Negri}]{Agrawal2018ContextualHI}
Ruchit~Rajeshkumar Agrawal, Marco Turchi, and Matteo Negri. 2018.
\newblock Contextual handling in neural machine translation: Look behind, ahead
  and on both sides.
\newblock In \emph{21st Annual Conference of the European Association for
  Machine Translation}, pages 11--20.

\bibitem[{Artetxe et~al.(2018)Artetxe, Labaka, Agirre, and
  Cho}]{artetxe2018unsupervised}
Mikel Artetxe, Gorka Labaka, Eneko Agirre, and Kyunghyun Cho. 2018.
\newblock \href {https://openreview.net/forum?id=Sy2ogebAW} {Unsupervised
  neural machine translation}.
\newblock In \emph{International Conference on Learning Representations}.

\bibitem[{Bahdanau et~al.(2015)Bahdanau, Cho, and Bengio}]{Bahdanau-2015-nmt}
Dzmitry Bahdanau, Kyunghyun Cho, and Yoshua Bengio. 2015.
\newblock \href {http://arxiv.org/abs/1409.0473} {Neural machine translation by
  jointly learning to align and translate}.
\newblock In \emph{3rd International Conference on Learning Representations,
  {ICLR} 2015, San Diego, CA, USA, May 7-9, 2015, Conference Track
  Proceedings}.

\bibitem[{Banerjee and Lavie(2005)}]{banerjee-lavie-2005-meteor}
Satanjeev Banerjee and Alon Lavie. 2005.
\newblock \href {https://aclanthology.org/W05-0909} {{METEOR}: An automatic
  metric for {MT} evaluation with improved correlation with human judgments}.
\newblock In \emph{Proceedings of the {ACL} Workshop on Intrinsic and Extrinsic
  Evaluation Measures for Machine Translation and/or Summarization}, pages
  65--72, Ann Arbor, Michigan. Association for Computational Linguistics.

\bibitem[{Ba{\~n}{\'o}n et~al.(2020)Ba{\~n}{\'o}n, Chen, Haddow, Heafield,
  Hoang, Espl{\`a}-Gomis, Forcada, Kamran, Kirefu, Koehn, Ortiz~Rojas,
  Pla~Sempere, Ram{\'\i}rez-S{\'a}nchez, Sarr{\'\i}as, Strelec, Thompson,
  Waites, Wiggins, and Zaragoza}]{banon-etal-2020-paracrawl}
Marta Ba{\~n}{\'o}n, Pinzhen Chen, Barry Haddow, Kenneth Heafield, Hieu Hoang,
  Miquel Espl{\`a}-Gomis, Mikel~L. Forcada, Amir Kamran, Faheem Kirefu, Philipp
  Koehn, Sergio Ortiz~Rojas, Leopoldo Pla~Sempere, Gema
  Ram{\'\i}rez-S{\'a}nchez, Elsa Sarr{\'\i}as, Marek Strelec, Brian Thompson,
  William Waites, Dion Wiggins, and Jaume Zaragoza. 2020.
\newblock \href {https://doi.org/10.18653/v1/2020.acl-main.417} {{P}ara{C}rawl:
  Web-scale acquisition of parallel corpora}.
\newblock In \emph{Proceedings of the 58th Annual Meeting of the Association
  for Computational Linguistics}, pages 4555--4567, Online. Association for
  Computational Linguistics.

\bibitem[{Belinkov and Bisk(2018)}]{belinkov2018synthetic}
Yonatan Belinkov and Yonatan Bisk. 2018.
\newblock \href {https://openreview.net/forum?id=BJ8vJebC-} {Synthetic and
  natural noise both break neural machine translation}.
\newblock In \emph{International Conference on Learning Representations}.

\bibitem[{Caswell(2020)}]{coswell-2020-googletrans}
Isaac Caswell. 2020.
\newblock \href
  {http://web.archive.org/web/20211225005922/https://blog.google/products/translate/five-new-languages/}
  {Google translate adds five languages}.
\newblock Accessed: 2022-01-14.

\bibitem[{Chen et~al.(2020)Chen, Liu, Chen, Jiang, and
  Liu}]{chen-etal-2020-accurate}
Yun Chen, Yang Liu, Guanhua Chen, Xin Jiang, and Qun Liu. 2020.
\newblock \href {https://doi.org/10.18653/v1/2020.emnlp-main.42} {Accurate word
  alignment induction from neural machine translation}.
\newblock In \emph{Proceedings of the 2020 Conference on Empirical Methods in
  Natural Language Processing (EMNLP)}, pages 566--576, Online. Association for
  Computational Linguistics.

\bibitem[{Cheng et~al.(2018)Cheng, Tu, Meng, Zhai, and
  Liu}]{cheng-etal-2018-towards}
Yong Cheng, Zhaopeng Tu, Fandong Meng, Junjie Zhai, and Yang Liu. 2018.
\newblock \href {https://doi.org/10.18653/v1/P18-1163} {Towards robust neural
  machine translation}.
\newblock In \emph{Proceedings of the 56th Annual Meeting of the Association
  for Computational Linguistics (Volume 1: Long Papers)}, pages 1756--1766,
  Melbourne, Australia. Association for Computational Linguistics.

\bibitem[{Clinchant et~al.(2019)Clinchant, Jung, and
  Nikoulina}]{clinchant-etal-2019-use}
Stephane Clinchant, Kweon~Woo Jung, and Vassilina Nikoulina. 2019.
\newblock \href {https://doi.org/10.18653/v1/D19-5611} {On the use of {BERT}
  for neural machine translation}.
\newblock In \emph{Proceedings of the 3rd Workshop on Neural Generation and
  Translation}, pages 108--117, Hong Kong. Association for Computational
  Linguistics.

\bibitem[{Dabre et~al.(2017)Dabre, Cromier{\`{e}}s, and
  Kurohashi}]{DBLP:journals/corr/DabreCK17}
Raj Dabre, Fabien Cromier{\`{e}}s, and Sadao Kurohashi. 2017.
\newblock \href {http://arxiv.org/abs/1702.06135} {Enabling multi-source neural
  machine translation by concatenating source sentences in multiple languages}.
\newblock \emph{CoRR}, abs/1702.06135.

\bibitem[{Doddington(2002)}]{doddington-2002-NIST}
George Doddington. 2002.
\newblock \href {https://dl.acm.org/doi/10.5555/1289189.1289273} {Automatic
  evaluation of machine translation quality using n-gram co-occurrence
  statistics}.
\newblock In \emph{Proceedings of the Second International Conference on Human
  Language Technology Research}, HLT '02, page 138–145, San Francisco, CA,
  USA. Morgan Kaufmann Publishers Inc.

\bibitem[{Garg et~al.(2019)Garg, Peitz, Nallasamy, and
  Paulik}]{garg-etal-2019-jointly}
Sarthak Garg, Stephan Peitz, Udhyakumar Nallasamy, and Matthias Paulik. 2019.
\newblock \href {https://doi.org/10.18653/v1/D19-1453} {Jointly learning to
  align and translate with transformer models}.
\newblock In \emph{Proceedings of the 2019 Conference on Empirical Methods in
  Natural Language Processing and the 9th International Joint Conference on
  Natural Language Processing (EMNLP-IJCNLP)}, pages 4453--4462, Hong Kong,
  China. Association for Computational Linguistics.

\bibitem[{Gowda and May(2020)}]{gowda-may-2020-finding}
Thamme Gowda and Jonathan May. 2020.
\newblock \href {https://doi.org/10.18653/v1/2020.findings-emnlp.352} {Finding
  the optimal vocabulary size for neural machine translation}.
\newblock In \emph{Findings of the Association for Computational Linguistics:
  EMNLP 2020}, pages 3955--3964, Online. Association for Computational
  Linguistics.

\bibitem[{Gowda et~al.(2021{\natexlab{a}})Gowda, You, Lignos, and
  May}]{gowda-etal-2021-macro}
Thamme Gowda, Weiqiu You, Constantine Lignos, and Jonathan May.
  2021{\natexlab{a}}.
\newblock \href {https://doi.org/10.18653/v1/2021.naacl-main.90}
  {Macro-average: Rare types are important too}.
\newblock In \emph{Proceedings of the 2021 Conference of the North American
  Chapter of the Association for Computational Linguistics: Human Language
  Technologies}, pages 1138--1157, Online. Association for Computational
  Linguistics.

\bibitem[{Gowda et~al.(2021{\natexlab{b}})Gowda, Zhang, Mattmann, and
  May}]{gowda-etal-2021-many}
Thamme Gowda, Zhao Zhang, Chris Mattmann, and Jonathan May. 2021{\natexlab{b}}.
\newblock \href {https://doi.org/10.18653/v1/2021.acl-demo.37}
  {Many-to-{E}nglish machine translation tools, data, and pretrained models}.
\newblock In \emph{Proceedings of the 59th Annual Meeting of the Association
  for Computational Linguistics and the 11th International Joint Conference on
  Natural Language Processing: System Demonstrations}, pages 306--316, Online.
  Association for Computational Linguistics.

\bibitem[{Gupta et~al.(2021)Gupta, Vavre, and
  Sarawagi}]{gupta-etal-2021-training}
Abhirut Gupta, Aditya Vavre, and Sunita Sarawagi. 2021.
\newblock \href {https://doi.org/10.18653/v1/2021.naacl-main.459} {Training
  data augmentation for code-mixed translation}.
\newblock In \emph{Proceedings of the 2021 Conference of the North American
  Chapter of the Association for Computational Linguistics: Human Language
  Technologies}, pages 5760--5766, Online. Association for Computational
  Linguistics.

\bibitem[{Johnson et~al.(2017)Johnson, Schuster, Le, Krikun, Wu, Chen, Thorat,
  Vi{\'e}gas, Wattenberg, Corrado, Hughes, and
  Dean}]{johnson-etal-2017-googles}
Melvin Johnson, Mike Schuster, Quoc~V. Le, Maxim Krikun, Yonghui Wu, Zhifeng
  Chen, Nikhil Thorat, Fernanda Vi{\'e}gas, Martin Wattenberg, Greg Corrado,
  Macduff Hughes, and Jeffrey Dean. 2017.
\newblock \href {https://doi.org/10.1162/tacl_a_00065} {{G}oogle{'}s
  multilingual neural machine translation system: Enabling zero-shot
  translation}.
\newblock \emph{Transactions of the Association for Computational Linguistics},
  5:339--351.

\bibitem[{Kondo et~al.(2021)Kondo, Hotate, Hirasawa, Kaneko, and
  Komachi}]{kondo-etal-2021-sentence}
Seiichiro Kondo, Kengo Hotate, Tosho Hirasawa, Masahiro Kaneko, and Mamoru
  Komachi. 2021.
\newblock \href {https://doi.org/10.18653/v1/2021.naacl-srw.18} {Sentence
  concatenation approach to data augmentation for neural machine translation}.
\newblock In \emph{Proceedings of the 2021 Conference of the North American
  Chapter of the Association for Computational Linguistics: Student Research
  Workshop}, pages 143--149, Online. Association for Computational Linguistics.

\bibitem[{Lample et~al.(2018)Lample, Conneau, Denoyer, and
  Ranzato}]{lample2018unsupervised}
Guillaume Lample, Alexis Conneau, Ludovic Denoyer, and Marc'Aurelio Ranzato.
  2018.
\newblock \href {https://openreview.net/forum?id=rkYTTf-AZ} {Unsupervised
  machine translation using monolingual corpora only}.
\newblock In \emph{International Conference on Learning Representations}.

\bibitem[{Mohan and Skotdal(2021)}]{mohan-2021-ms-translator}
Krishna~Doss Mohan and Jann Skotdal. 2021.
\newblock \href
  {http://web.archive.org/web/20211203095101/https://www.microsoft.com/en-us/research/blog/microsoft-translator-now-translating-100-languages-and-counting/}
  {Microsoft translator: Now translating 100 languages and counting!}
\newblock Accessed: 2022-01-14.

\bibitem[{M{\"u}ller et~al.(2020)M{\"u}ller, Rios, and
  Sennrich}]{muller-etal-2020-domain}
Mathias M{\"u}ller, Annette Rios, and Rico Sennrich. 2020.
\newblock \href {https://aclanthology.org/2020.amta-research.14} {Domain
  robustness in neural machine translation}.
\newblock In \emph{Proceedings of the 14th Conference of the Association for
  Machine Translation in the Americas (Volume 1: Research Track)}, pages
  151--164, Virtual. Association for Machine Translation in the Americas.

\bibitem[{Myers-Scotton(1989)}]{myers1989codeswitching}
Carol Myers-Scotton. 1989.
\newblock Codeswitching with english: types of switching, types of communities.
\newblock \emph{World Englishes}, 8(3):333--346.

\bibitem[{Myers-Scotton and Ury(1977)}]{cms-and-ury-1977-biling}
Carol Myers-Scotton and William Ury. 1977.
\newblock \href {https://doi.org/doi:10.1515/ijsl.1977.13.5} {Bilingual
  strategies: The social functions of code-switching}.
\newblock \emph{Linguistics: An Interdisciplinary Journal of the Language
  Sciences}, 1977(13):5--20.

\bibitem[{Nakazawa et~al.(2021)Nakazawa, Nakayama, Ding, Dabre, Higashiyama,
  Mino, Goto, Pa~Pa, Kunchukuttan, Parida, Bojar, Chu, Eriguchi, Abe, Oda, and
  Kurohashi}]{nakazawa-etal-2021-overview}
Toshiaki Nakazawa, Hideki Nakayama, Chenchen Ding, Raj Dabre, Shohei
  Higashiyama, Hideya Mino, Isao Goto, Win Pa~Pa, Anoop Kunchukuttan,
  Shantipriya Parida, Ond{\v{r}}ej Bojar, Chenhui Chu, Akiko Eriguchi, Kaori
  Abe, Yusuke Oda, and Sadao Kurohashi. 2021.
\newblock \href {https://doi.org/10.18653/v1/2021.wat-1.1} {Overview of the 8th
  workshop on {A}sian translation}.
\newblock In \emph{Proceedings of the 8th Workshop on Asian Translation
  (WAT2021)}, pages 1--45, Online. Association for Computational Linguistics.

\bibitem[{Nguyen et~al.(2021)Nguyen, Murray, and
  Chiang}]{nguyen-etal-2021-data}
Toan~Q. Nguyen, Kenton Murray, and David Chiang. 2021.
\newblock \href {https://doi.org/10.18653/v1/2021.iwslt-1.33} {Data
  augmentation by concatenation for low-resource translation: A mystery and a
  solution}.
\newblock In \emph{Proceedings of the 18th International Conference on Spoken
  Language Translation (IWSLT 2021)}, pages 287--293, Bangkok, Thailand
  (online). Association for Computational Linguistics.

\bibitem[{Nilep(2006)}]{nilep-2006-codeswitch}
Chad Nilep. 2006.
\newblock \href {https://doi.org/10.25810/hnq4-jv62} {“code switching” in
  sociocultural linguistics}.
\newblock \emph{Colorado Research in Linguistics}, 19.

\bibitem[{Niu et~al.(2020)Niu, Mathur, Dinu, and
  Al-Onaizan}]{niu-etal-2020-evaluating}
Xing Niu, Prashant Mathur, Georgiana Dinu, and Yaser Al-Onaizan. 2020.
\newblock \href {https://doi.org/10.18653/v1/2020.acl-main.755} {Evaluating
  robustness to input perturbations for neural machine translation}.
\newblock In \emph{Proceedings of the 58th Annual Meeting of the Association
  for Computational Linguistics}, pages 8538--8544, Online. Association for
  Computational Linguistics.

\bibitem[{Och and Ney(2001)}]{och-ney-2001-statistical}
Franz~Josef Och and Hermann Ney. 2001.
\newblock \href {https://aclanthology.org/2001.mtsummit-papers.46} {Statistical
  multi-source translation}.
\newblock In \emph{Proceedings of Machine Translation Summit VIII}, Santiago de
  Compostela, Spain.

\bibitem[{Popovi{\'c}(2015)}]{popovic-2015-chrf}
Maja Popovi{\'c}. 2015.
\newblock \href {https://doi.org/10.18653/v1/W15-3049} {chr{F}: character
  n-gram {F}-score for automatic {MT} evaluation}.
\newblock In \emph{Proceedings of the Tenth Workshop on Statistical Machine
  Translation}, pages 392--395, Lisbon, Portugal. Association for Computational
  Linguistics.

\bibitem[{Ribeiro et~al.(2020)Ribeiro, Wu, Guestrin, and
  Singh}]{ribeiro-etal-2020-beyond}
Marco~Tulio Ribeiro, Tongshuang Wu, Carlos Guestrin, and Sameer Singh. 2020.
\newblock \href {https://doi.org/10.18653/v1/2020.acl-main.442} {Beyond
  accuracy: Behavioral testing of {NLP} models with {C}heck{L}ist}.
\newblock In \emph{Proceedings of the 58th Annual Meeting of the Association
  for Computational Linguistics}, pages 4902--4912, Online. Association for
  Computational Linguistics.

\bibitem[{Sennrich et~al.(2016{\natexlab{a}})Sennrich, Haddow, and
  Birch}]{sennrich-etal-2016-improving}
Rico Sennrich, Barry Haddow, and Alexandra Birch. 2016{\natexlab{a}}.
\newblock \href {https://doi.org/10.18653/v1/P16-1009} {Improving neural
  machine translation models with monolingual data}.
\newblock In \emph{Proceedings of the 54th Annual Meeting of the Association
  for Computational Linguistics (Volume 1: Long Papers)}, pages 86--96, Berlin,
  Germany. Association for Computational Linguistics.

\bibitem[{Sennrich et~al.(2016{\natexlab{b}})Sennrich, Haddow, and
  Birch}]{sennrich-etal-2016-neural}
Rico Sennrich, Barry Haddow, and Alexandra Birch. 2016{\natexlab{b}}.
\newblock \href {https://doi.org/10.18653/v1/P16-1162} {Neural machine
  translation of rare words with subword units}.
\newblock In \emph{Proceedings of the 54th Annual Meeting of the Association
  for Computational Linguistics (Volume 1: Long Papers)}, pages 1715--1725,
  Berlin, Germany. Association for Computational Linguistics.

\bibitem[{Snover et~al.(2006)Snover, Dorr, Schwartz, Micciulla, and
  Makhoul}]{snover-etal-2006-study}
Matthew Snover, Bonnie Dorr, Rich Schwartz, Linnea Micciulla, and John Makhoul.
  2006.
\newblock \href {https://aclanthology.org/2006.amta-papers.25} {A study of
  translation edit rate with targeted human annotation}.
\newblock In \emph{Proceedings of the 7th Conference of the Association for
  Machine Translation in the Americas: Technical Papers}, pages 223--231,
  Cambridge, Massachusetts, USA. Association for Machine Translation in the
  Americas.

\bibitem[{Song et~al.(2019)Song, Zhang, Yu, Luo, Wang, and
  Zhang}]{song-etal-2019-code}
Kai Song, Yue Zhang, Heng Yu, Weihua Luo, Kun Wang, and Min Zhang. 2019.
\newblock \href {https://doi.org/10.18653/v1/N19-1044} {Code-switching for
  enhancing {NMT} with pre-specified translation}.
\newblock In \emph{Proceedings of the 2019 Conference of the North {A}merican
  Chapter of the Association for Computational Linguistics: Human Language
  Technologies, Volume 1 (Long and Short Papers)}, pages 449--459, Minneapolis,
  Minnesota. Association for Computational Linguistics.

\bibitem[{Sutskever et~al.(2014)Sutskever, Vinyals, and
  Le}]{sutskever2014sequence}
Ilya Sutskever, Oriol Vinyals, and Quoc~V Le. 2014.
\newblock \href
  {https://proceedings.neurips.cc/paper/2014/file/a14ac55a4f27472c5d894ec1c3c743d2-Paper.pdf}
  {Sequence to sequence learning with neural networks}.
\newblock In \emph{Advances in Neural Information Processing Systems},
  volume~27. Curran Associates, Inc.

\bibitem[{Tiedemann(2020)}]{tiedemann-2020-tatoeba}
J{\"o}rg Tiedemann. 2020.
\newblock \href {https://aclanthology.org/2020.wmt-1.139} {The tatoeba
  translation challenge {--} realistic data sets for low resource and
  multilingual {MT}}.
\newblock In \emph{Proceedings of the Fifth Conference on Machine Translation},
  pages 1174--1182, Online. Association for Computational Linguistics.

\bibitem[{Tiedemann and Scherrer(2017)}]{tiedemann-scherrer-2017-neural}
J{\"o}rg Tiedemann and Yves Scherrer. 2017.
\newblock \href {https://doi.org/10.18653/v1/W17-4811} {Neural machine
  translation with extended context}.
\newblock In \emph{Proceedings of the Third Workshop on Discourse in Machine
  Translation}, pages 82--92, Copenhagen, Denmark. Association for
  Computational Linguistics.

\bibitem[{Vaibhav et~al.(2019)Vaibhav, Singh, Stewart, and
  Neubig}]{vaibhav-etal-2019-improving}
Vaibhav Vaibhav, Sumeet Singh, Craig Stewart, and Graham Neubig. 2019.
\newblock \href {https://doi.org/10.18653/v1/N19-1190} {Improving robustness of
  machine translation with synthetic noise}.
\newblock In \emph{Proceedings of the 2019 Conference of the North {A}merican
  Chapter of the Association for Computational Linguistics: Human Language
  Technologies, Volume 1 (Long and Short Papers)}, pages 1916--1920,
  Minneapolis, Minnesota. Association for Computational Linguistics.

\bibitem[{Vaswani et~al.(2017)Vaswani, Shazeer, Parmar, Uszkoreit, Jones,
  Gomez, Kaiser, and Polosukhin}]{vaswani-2017-attention}
Ashish Vaswani, Noam Shazeer, Niki Parmar, Jakob Uszkoreit, Llion Jones,
  Aidan~N Gomez, {\L}ukasz Kaiser, and Illia Polosukhin. 2017.
\newblock \href
  {https://proceedings.neurips.cc/paper/2017/file/3f5ee243547dee91fbd053c1c4a845aa-Paper.pdf}
  {Attention is all you need}.
\newblock In \emph{Advances in Neural Information Processing Systems},
  volume~30. Curran Associates, Inc.

\bibitem[{Wu et~al.(2016)Wu, Schuster, Chen, Le, Norouzi, Macherey, Krikun,
  Cao, Gao, Macherey et~al.}]{wu2016google}
Yonghui Wu, Mike Schuster, Zhifeng Chen, Quoc~V Le, Mohammad Norouzi, Wolfgang
  Macherey, Maxim Krikun, Yuan Cao, Qin Gao, Klaus Macherey, et~al. 2016.
\newblock \href {https://arxiv.org/abs/1609.08144} {Google's neural machine
  translation system: Bridging the gap between human and machine translation}.
\newblock \emph{arXiv preprint arXiv:1609.08144}.

\bibitem[{Yang et~al.(2020)Yang, Hu, Han, Huang, and Ju}]{yang-etal-2020-csp}
Zhen Yang, Bojie Hu, Ambyera Han, Shen Huang, and Qi~Ju. 2020.
\newblock \href {https://doi.org/10.18653/v1/2020.emnlp-main.208}
  {{CSP}:code-switching pre-training for neural machine translation}.
\newblock In \emph{Proceedings of the 2020 Conference on Empirical Methods in
  Natural Language Processing (EMNLP)}, pages 2624--2636, Online. Association
  for Computational Linguistics.

\bibitem[{Zenkel et~al.(2019)Zenkel, Wuebker, and
  DeNero}]{DBLP:journals/corr/abs-1901-11359}
Thomas Zenkel, Joern Wuebker, and John DeNero. 2019.
\newblock \href {http://arxiv.org/abs/1901.11359} {Adding interpretable
  attention to neural translation models improves word alignment}.
\newblock \emph{CoRR}, abs/1901.11359.

\bibitem[{Zenkel et~al.(2020)Zenkel, Wuebker, and
  DeNero}]{zenkel-etal-2020-end}
Thomas Zenkel, Joern Wuebker, and John DeNero. 2020.
\newblock \href {https://doi.org/10.18653/v1/2020.acl-main.146} {End-to-end
  neural word alignment outperforms {GIZA}++}.
\newblock In \emph{Proceedings of the 58th Annual Meeting of the Association
  for Computational Linguistics}, pages 1605--1617, Online. Association for
  Computational Linguistics.

\bibitem[{Zhang et~al.(2020)Zhang, Williams, Titov, and
  Sennrich}]{zhang-etal-2020-improving}
Biao Zhang, Philip Williams, Ivan Titov, and Rico Sennrich. 2020.
\newblock \href {https://doi.org/10.18653/v1/2020.acl-main.148} {Improving
  massively multilingual neural machine translation and zero-shot translation}.
\newblock In \emph{Proceedings of the 58th Annual Meeting of the Association
  for Computational Linguistics}, pages 1628--1639, Online. Association for
  Computational Linguistics.

\end{thebibliography}
\balance

\end{document}